\documentclass{article}

\usepackage{subcaption}
\usepackage{arxiv}
\usepackage{listings}
\usepackage[utf8]{inputenc} 
\usepackage[T1]{fontenc}    
\usepackage{hyperref}       
\usepackage{url}            
\usepackage{booktabs}       
\usepackage{amsfonts}       
\usepackage{nicefrac}       
\usepackage{microtype}      
\usepackage{lipsum}		
\usepackage{graphicx}
\usepackage[square,sort,comma,numbers]{natbib}
\usepackage{doi}
\usepackage{tcolorbox}
\usepackage{xcolor}
\usepackage[linesnumbered,ruled,vlined]{algorithm2e}
\usepackage{tcolorbox}

\SetCommentSty{mycommfont}

\SetKwInput{KwInput}{Input}                
\SetKwInput{KwOutput}{Output}              

\title{Y Social: an LLM-powered Social Media Digital Twin}

\date{} 					

\author{Giulio Rossetti\\
	CNR-ISTI, Italy\\
	\texttt{giulio.rossetti@isti.cnr.it} \\
	\And
	Massimo Stella \\
	University of Trento, Italy\\
	\texttt{massimo.stella-1@unitn.it} \\
 \And
	Rémy Cazabet \\
	University of Lyon, France\\
	\texttt{remy.cazabet@univ-lyon1.fr} \\
 \And
	Katherine Abramski \\
	University of Pisa, Italy \\
	 \texttt{katherine.abramski@phd.unipi.it} \\
     \And
	Erica Cau \\
	University of Pisa, Italy \\
	 \texttt{erica.cau@phd.unipi.it} \\
	 \And
	 Salvatore Citraro \\
      CNR-ISTI, Italy\\
	 \texttt{salvatore.citraro@isti.cnr.it} \\
	 \And
	 Andrea Failla \\
	 University of Pisa, Italy \\
	 \texttt{andrea.failla@phd.unipi.it} \\
        \And
  Riccardo Improta \\
	University of Trento, Italy\\
	 \texttt{riccardo.improta@unitn.it} \\
  \And
  Virginia Morini \\
	 University of Pisa, Italy \\
	 \texttt{virginia.morini@phd.unipi.it} \\
  \And
  Valentina Pansanella \\
	 CNR-ISTI, Italy\\
	 \texttt{valentina.pansanella@isti.cnr.it} \\
}

\renewcommand{\shorttitle}{Y: Where the Digital World Comes to Life}

\hypersetup{
pdftitle={A template for the arxiv style},
pdfsubject={q-bio.NC, q-bio.QM},
pdfauthor={Giulio Rossetti},
pdfkeywords={First keyword, Second keyword, More},
}

\begin{document}
\maketitle
\begin{abstract}
In this paper we introduce {\tt Y}, a new-generation digital twin designed to replicate an online social media platform. 
Digital twins are virtual replicas of physical systems that allow for advanced analyses and experimentation. In the case of social media, a digital twin such as {\tt Y} provides a powerful tool for researchers to simulate and understand complex online interactions. 
{\tt Y} leverages state-of-the-art Large Language Models (LLMs) to replicate sophisticated agent behaviors, enabling accurate simulations of user interactions, content dissemination, and network dynamics. 
By integrating these aspects, {\tt Y} offers valuable insights into user engagement, information spread, and the impact of platform policies. 
Moreover, the integration of LLMs allows {\tt Y} to generate nuanced textual content and predict user responses, facilitating the study of emergent phenomena in online environments.

To better characterize the proposed digital twin, in this paper we describe the rationale behind its implementation, provide examples of the analyses that can be performed on the data it enables to be generated, and discuss its relevance for multidisciplinary research.
\end{abstract}


\section{Introduction}
\label{sec:intro}
Online social media (OSM henceforth) have revolutionized the way we exchange information. 
From the user's perspective, these digital ecosystems are largely effortless \cite{siddiqui2016social}, enabling convenient ways of exchanging personal content \cite{abramski2024voices}, seeking information \cite{santoro2023analyzing} and synchronizing with others \cite{david2012cooperation}. 
This convenience has catalyzed a massive digital shift in social and information exchanges from offline to online settings \cite{siddiqui2016social}, which has provided novel access to massive amounts of online data regarding human behaviour \cite{stella2023using}. 
Unconstrained by geographical barriers, the massive adoption of social media has given rise to novel phenomena that are absent in in-person interactions, such as the influence of complexity and artificial intelligence. 
Complexity in social media is strongly related to the motto "more is different" \cite{anderson1972more}: the idea that the co-occurrence of many, even similar, interactions within the same context can lead to unexpected phenomena. 
Examples include acts as simple and seemingly insignificant as following another user, or re-sharing content. 
Taken individually, these actions can be understood in terms of a user's activity, psychology, and engagement \cite{milli2015quantification,stella2023using,morini2021toward}, but when repeated by vast amounts of users, these actions can determine the unexpected rise and fall of a massively followed account (influencer) or they can create cascades of information re-sharing (viral content) \cite{siddiqui2016social}. Artificial intelligences (AIs) also influence online interactions since social media actions such as following, re-sharing, posting, and commenting can nowadays also be performed by non-human users. 
Social bots are AIs that pose as humans \cite{ferrara2016rise} and interact with human users, either beneficially or maliciously \cite{stella2018bots}. 
For instance, social bots can contribute positively to social media pluralism by automatically diffusing news and links that are external to social media platforms \cite{stella2019influence,varol2020journalists}. 
However, social bots can also act by broadcasting biased content or diffusing misinformation, which exposes human users to a wider variety of harmful content \cite{stella2018bots,cresci2020decade}.

Despite existing online, content engagement on social media can deeply affect humans in their real-world surroundings \cite{kramer2014experimental,rosen2013facebook}. 
Besides acting as a distraction \cite{rosen2013facebook}, online posts can influence the emotions experienced by an individual, a phenomenon known as emotional contagion, which has been identified on platforms like Facebook \cite{kramer2014experimental}. 
Furthermore, social media interactions are mainly textual in nature, which can lead to some unresolved ambiguity in understanding communicative intents \cite{stella2023using}. 
Dissonance is further exacerbated by the fact that social media content is often rich in cognitive, emotional, and psychological features \cite{stella2023using,chandrasekaran2021multimodal}. 
For instance, a post that expresses ideas in words or hashtags can also contain a familiar picture or emoticons that resemble recognizable facial expressions \cite{stella2018bots}.
The negative impact of unresolved ambiguity and toxic or dissonant content in social media on mental health remains poorly understood \cite{berryman2018social}. 
While a causal relationship has yet to be determined, several recent studies have begun to link the longer hours spent on social media to depression disorders \cite{braghieri2022social,berryman2018social}.

Social media interactions might not depend solely on users themselves \cite{morini2024perils}. 
In contrast to in-person interactions, the experiences of online users are often reinforced by platform-specific algorithmic curation \cite{fabbri2022exposure}, i.e. embedding recommender or ranking systems in social media for bolstering user engagement. 
Recommender and ranking systems are algorithms that constantly suggest or re-direct engagement towards specific users or content. 
These algorithms ultimately guide the attention of vast audiences towards specific trending topics, influencers, or conversations with leanings similar to one's own \cite{cinus2022effect,brown2022echo}. 
Algorithmic curation can reinforce users' biases while preventing healthy discussions, thus contributing to users' fragmentation across cognitive or social groups \cite{hosanagar2014will,morini2021toward}.
These aspects raise crucial technical and ethical questions about balancing possibilities and risks concerning the role of human-AI interactions on social dynamics regarding information diffusion, opinion formation, and mental health \cite{morini2024perils,pedreschi2023human}. 

{\bf Why does having a Digital Twin of online social media platforms make sense?}

OSMs are complex systems characterized by countless emerging behaviors - both at the structural and social levels.
Data-driven social studies provide valuable insights into such complexity and the phenomena it might generate.
However, they suffer from natural limitations that often reduce their potential impact.
Particularly, such studies often rely on limited - nonlongitudinal - data samples, cannot explicitly account for the ``ìnvisible" although impactful hands of algorithmic curation, and are prone to noise due to the not always measurable impact the selected data sampling strategy has on the observed results. 
Moreover, they are dependent on the data access provided by online social media platforms (e.g., X/Twitter, Meta's Facebook/Instagram/Threads, Reddit, and others): access that is reducing over time even for research purposes - notwithstanding national and international efforts to regulate it (e.g., through the EU's Digital Service Act\footnote{EU DSA: \url{https://shorturl.at/nLj0v}}).

All such considerations vouch for identifying complementary data sources that can offer completeness and algorithmic control and are not tied to access policies set by third-party entities.
Online Social Media digital twins offer a first step in such a direction: they allow researchers to formulate hypotheses and test them in OSM-like controlled environments - accounting for such variables that open-ended data studies cannot model/infer.
Indeed, such tools - supported by recent advances in Artificial Intelligence - cannot substitute data-driven studies; however, they can support in-vitro experiments fitted on real-world data and allow for a better understanding of the phenomena observed in the wild.

{\bf Which phenomena can emerge as a result of OSM simulated environments powered by LLM agents?}

The rapid evolution of AI \cite{pedreschi2023human} has produced novel gaps in understanding last-generation social bots, powered by Large Language Models (LLMs) \cite{achiam2023gpt, llama3modelcard}. 
LLMs are generative AI systems capable of understanding and generating human-like text even within social media ecosystems. 
They can generate convincing human texts with minimal human instructions -- just simple prompts -- without syntactic or grammatical errors \cite{achiam2023gpt,abramski2023cognitive}. 
They can even impersonate specific profiles \cite{de2024introducing}, e.g. posing as a statistics expert with a right-wing political leaning and high creativity, or express nuanced human emotions \cite{jiang2024evaluating}. 
Such human-like capabilities are enabled by the fact that LLMs are trained on billions and billions of human-generated text.

Exploring LLMs' role within social media remains a crucial yet largely unexplored research question. 
While research on human-LLM interactions in social media considerably overlaps with past studies about humans and social bots \cite{ferrara2016rise,cresci2020decade}, the current research literature leaves a more interesting and unique research gap.
Digital twins are reference models, simulated to gain insights about a physical counterpart \cite{schluse2018experimentable}. 
For instance, an LLM might impersonate a patient in clinical therapy and thus be considered its digital twin \cite{de2024introducing}. 
Adopting LLMs as synthetic agents/online users can crucially proxy human behaviour, in line with several upcoming studies exploring "artificial humans" as digital twins of real-world users \cite{de2023emergence,de2024introducing}. 
Milestone research suggests that populations of these models can exhibit emergent behaviors similar to those found in humans, such as scale-free networks \cite{de2023emergence}, patterns of information diffusion \cite{gao2023s} and affective biases \cite{abramski2023cognitive}. 
These preliminary results offer new opportunities for developing controlled environments in which researchers can gain a scientifically sound and interpretable understanding of real-world complex systems by investigating which empirical phenomena also arise in digital twins endowed with a limited set of rules.

{\bf Y: Where the Digital World Comes to Life.} In view of the above research gap and the promising opportunities opened up by LLMs, in this work we introduce {\tt Y}, a social media platform powered by LLMs, serving as a digital twin for complex agent-based social simulations.

\begin{tcolorbox}[title={So, why Y?}]
Developing a platform like {\tt Y} has several implications and motivations. 
\begin{itemize}
    \item Firstly, by observing how LLMs interact with each other in a social media setting, researchers can gain insights into AI behavior, biases and cognitive capabilities within simulated social contexts. 
    \item 
    Secondly, enabling algorithmic curation in a digital twin of social systems makes it possible to study how recommendation systems can impact user behaviors. 
    \item Last but not least, using digital twins to simulate social media interactions via LLMs sets the stage for unprecedented experiments that can test the cognitive, psychological, social, and communicative facets of real social media.
\end{itemize}

\end{tcolorbox}

In the subsequent sections, we first review the recent research literature that has been pivotal in the development of {\tt Y} (Section \ref{sec:related}). Following this, we present a detailed technical overview of the system (Section \ref{sec:Y}) and illustrate its capabilities through a simple case study (Section \ref{sec:case}) to provide better context for its behaviors.
Finally, we explore the potential impacts of a social media digital twin on multidisciplinary computational social science research (Section \ref{sec:multid}), and conclude with a brief discussion on the future evolution of the project (Section \ref{sec:conclusions}).
\newpage
\section{Related Works}
\label{sec:related}
This section gives an overview of the relevant literature concerning the development of {\tt Y}. 
We consider research on Online Social Platforms (OSP) as either simulation studies or empirical studies, following the highest level of the taxonomy defined in \cite{pappalardo2024survey}. 
Focusing on simulation studies, we delve into agent-based simulations, discussing how OSPs are integrated, and we discuss recent advancements since the advent of LLMs, considering the move towards creating OSP ``digital twins". 
We also provide an overview of some of the main tools for social simulations.

{\bf Computational Social Science.} Collections of things can have properties that don't apply to their individual parts. 
For example, the property of pressure pertains to gases as a whole, resulting from the interactions among molecules. 
The same idea that applies in physical substances also applies to societies, though humans are far more complex than molecules. 
The field of \textit{computational social science} \cite{conte2012manifesto} — at the crossroads of complex systems, network science, and big data analytics — seeks to discover universal laws of human behavior through computational methods. 
In the natural sciences, experimental designs allow stringent control over variables, enabling precise isolation and examination of causal relationships. 
This control aids in replicating experiments and validating results. 
On the contrary, in the social sciences, experiments are challenging due to both practical and ethical constraints. 
Fortunately, the advent of the digital age has not only led to the production of an unprecedented amount of data traces from human activities, but enhancements in computational means has also opened possibilities for more complex experimental setups, overcoming previous difficulties. 
Social sciences studies utilize a variety of methods, such as surveys \cite{bowling2005quantitative}, computer simulations \cite{gilbert2005simulation}, observational research -- encompassing statistical analysis \cite{mukherjee2018statistical}, network analysis \cite{wasserman1994social} and data mining \cite{attewell2015data} --, and experimental and quasi-experimental approaches \cite{shadish2002quasi, jackson2013principles}.

Defined at a high level, social simulations \cite{squazzoni2014social} are computational models designed to replicate the behavior and interactions of individuals or groups within a social system. 
Simulations can be implemented using different frameworks, Agent-Based Modeling \cite{bianchi2015agent} being one of the most commonly employed. 
In any case humans -- and other individual elements, such as institutions, technologies, etc. -- can be seen as software entities called \textit{agents} which can be part of an \textit{environment} and can be embedded in a \textit{social network} \cite{granovetter1973strength} influencing their decisions and behaviours. 
As the latter is often represented as a graph -- being fully connected \cite{sirbu2019algorithmic} or having a more complex configuration \cite{pansanella2022mean} --, agents can have a set of edges, representing their social relationships, which can be static or dynamics and may hold attributes representing characteristics of the relationship, e.g., strength, trust, or influence.
For example, it may be that the network of interactions evolves as a simulation progresses as a consequence of a pre-defined rule of the model, as in the Barabasi-Albert model \cite{barabasi2003scale} of network formation, or of the evolution of other agent's attributes, as in adaptive networks models \cite{pansanella2022modeling}. 
Agents can, in fact, be enriched with \textit{attributes}, such as demographic factors or cognitive biases \cite{deffuant2000mixing}, which are mainly static during a single simulation and with \textit{states}, i.e., dynamic attributes dependent on social influence, representing factors like infection \cite{lorig2021agent}, adoption \cite{serrano2016validating}, opinions \cite{deffuant2000mixing}, or emotions \cite{schweitzer2010agent}, generally modelled as simple numerical variables and often the core element of the social simulation study. 
To execute the social simulation, each agent follows a set of \textit{rules}, which are algorithms dictating when an agent becomes active, the actions they can take, and the impact of these actions on themselves and the population, which can be more or less strict, giving the agent different degrees of autonomy.  
This flexible framework allows for high degrees of heterogeneity across different dimensions, creating agents with different beliefs, preferences, behaviours, values and positions in the social systems \cite{squazzoni2014social}. 
Social \emph{simulation} studies has been used in different fields to explore and understand emergent behaviours \cite{gilbert2006emergence} in social systems since the 1960s, with prominent examples from Conway's Game of Life \cite{conway1970game}, to the Schelling model of segregation \cite{schelling1971dynamic}, and Axelrod's prisoner dilemma \cite{axelrod1980effective}. 
More recently, social simulations have been used to study epidemics \cite{lorig2021agent}, diffusion of information \cite{guille2013information}, belief and emotion dynamics \cite{deffuant2000mixing}, economics and finance \cite{axtell2022agent}, urban dynamics \cite{cornacchia2021sts}, political sciences \cite{epstein2002modeling}, and others. 

{\bf Simulations of online platforms.} 
In recent years a great deal of attention has been dedicated to the study of such social dynamics within socio-technical systems \cite{vespignani2012modelling}.  
Many simulation studies try to explain online-specific phenomena, such as polarization, echo chambers or misinformation diffusion, with simple social characteristics (e.g. bounded confidence, homophily), without incorporating any specific characteristic of online environments \cite{quattrociocchi2014opinion,del2015echo,gargiulo2016role,del2017modeling,baumann2020modeling,chen2021opinion,peralta2024multidimensional}. 
Online-platforms are often explicitly modelled by incorporating recommender systems into the simulation. 
In several studies, a recommender system is used as a parameter in simulations to modify the probability of agent interactions \cite{sirbu2019algorithmic, pansanella2022mean, pansanella2022modeling, pansanella2023mass, valensise2023drivers, peralta2021effect, peralta2021opinion, cinus2022effect}, to alter the ordering of news feed posts \cite{perra2019modelling, gausen2022using, rossi2021closed, ribeiro2023amplification}, or to create new links in the network \cite{chitra2020analyzing}
Some research has incorporated state-of-the-art recommender systems \cite{jiang2019degenerate}, e.g., collaborative-filtering algorithms to suggest content or to suggest new links \cite{fabbri2022exposure, ferrara2022link}. 
However, characteristics of Online Social Platforms are not limited to their algorithms.    
Other works simulate online social platforms by incorporating specific behaviors like posting, retweeting, and user sessions, often tailored to specific platforms \cite{nasrinpour2016agent}.
Simulations also make use of different data from online social platforms \cite{pansanella2023mass, valensise2023drivers, monti2020learning} to calibrate micro-behaviours and/or validate macro-results.

\paragraph{LLM-enhanced Social Simulations}
Recent advancements have introduced \textit{generative agents} that leverage generative models -- especially language models \cite{brown2020language} -- to simulate realistic human behavior, achieving believable individual and collective behaviors \cite{park2023simulacra}. 
Among these, LLMs have garnered significant attention for their ability to produce emergent behaviors akin to human societies \cite{leng2023llm}.
The use of this framework is in its infancy and thus we see the use of very different methodologies and implementations that make it difficult to compare the results obtained, in the same way as in classical agent-based models. 
The number of different LLMs tested ranges from 1 \cite{park2023simulacra, park2022social, leng2023llm, de2023emergence, mou2024unveiling, chuang2023simulating, breum2023persuasive, li2023large}, to 2 or a few, \cite{shu2024llm, xie2024can, flamino2024limits, tornberg2023simulating}, to 10 or more \cite{la2024open, piatti2024cooperate}. 
Tested models vary from closed-source, such as the GPT series \cite{park2022social, park2023generative, park2023simulacra, leng2023llm, de2023emergence, mou2024unveiling, chuang2023simulating, tornberg2018echo, li2023large}, to open-sources, such as Llama, Mistral and others \cite{la2024open, breum2023persuasive}, or a mixture of the two \cite{shu2024llm, xie2024can, flamino2024limits, piatti2024cooperate}. 
In general, the aim of these latter studies is to compare results across different models. 
In these studies each agent is not a simple "software" following a pre-defined algorithm, but an instance of an LLM, interacting with humans, other agents and/or the environment through textual means. 
Exceptions create mixed populations of human and LLMs \cite{park2023simulacra, leng2023llm} or classical agents and LLM-agents \cite{mou2024unveiling}
Agents are often assigned different "personas" -- name, age, gender, different interests, personality traits, political leanings --, through textual prompts. 
Besides manually assigning different personas, in  \cite{he2024community} authors demonstrate the ability of fine-tuned LLMs agents to align with beliefs and opinions of different online communities (such as Subreddits) and replicate human responses in surveys. 
The implementation of a "memory module", that the agent can query to make decisions and take future actions is present in several studies \cite{park2023simulacra, mou2024unveiling, chuang2023simulating,breum2023persuasive, li2023large}.
Environments can be absent \cite{la2024open, leng2023llm, de2023emergence, xie2024can} -- with just a population of agents interacting with humans or with each other --, but also very complex. 
For example, in \cite{park2023simulacra}, authors created Smallville, a sandbox environment resembling a small village where agents are represented similarly to characters of The Sims. 
In \cite{park2022social}, agents interacted within SimReddit, a web-based prototyping tool designed for those who need to implement systems like Reddit. 
In \cite{mou2024unveiling} there is a Twitter-like environment with timelines where LLM-agents generated content is published and can be read by other agents. 
Authors in \cite{tornberg2023simulating} implement a general social media platform with different possible timeline curation algorithms. 
In \cite{park2023simulacra,de2023emergence} authors studied whether generative agents were able of information diffusion, relationship formation and coordination, while in \cite{park2022social} they simulated participation in online communities. 
In \cite{leng2023llm}, authors examined whether LLMs can display social learning, preference and indirect reciprocity.
Some studies replicate classical opinion evolution simulations, testing one or multiple models \cite{mou2024unveiling, chuang2023simulating, breum2023persuasive, flamino2024limits}. 
These recent studies also focus on understanding the interplay between such social dynamics and language \cite{breum2023persuasive}, e.g. levels of toxicity in online discussions \cite{tornberg2023simulating}, which was unfeasible exploiting classical agents. 
Without considering collective dynamics, some studies focus on assessing individual characteristics of generative agents.
In \cite{la2024open}, for example, LLM-agents performed multiple rounds of MBTI tests. 
Generative agents show human-like collective behaviours such as the formation of scale-free networks \cite{de2013simulation} and information diffusion \cite{gao2023s}. 
Moreover, these enhanced models outperform traditional ones in replicating echo-chambers dynamics in Twitter-like environments \cite{mou2024unveiling}, while showing consistent results on the role of confirmation bias as a cause of polarization and fragmentation \cite{chuang2023simulating}. 
In absence of possible comparisons with classical agents, generative agents showed the ability to generate persuasive arguments, in line with psycho-linguistic theories of opinion change \cite{monti2022language, breum2023persuasive}, and less toxic behaviours, being more polite and respectful than real social media users \cite{tornberg2023simulating}.
Similarities and differences of such agents with humans and of these simulations to real social systems need careful consideration to avoid oversimplified conclusions. 

Despite having advanced social simulations with generative agents, we are still far from creating a \textit{social platform digital twin}. 
Digital-Twins (DT) are defined as physical/virtual machines or computer-based models that are simulating or "twinning" a physical entity, continuously predicting future statuses and allowing simulations and testing \cite{barricelli2019survey}. 
Digital-Twins are already vastly employed in different fields from engineering \cite{tao2018digital} to healthcare \cite{sun2023digital}, becoming one of the top 10 strategic technology trends for 2019 in Gartner's list\footnote{\url{https://www.gartner.com/en/documents/3904569}}. 
Relatively new directions involve, but are not limited to, fields like sports \cite{hlis2024digital} and city science and urban computing \cite{xu2023urban, pereira2023smart, yossef2023social}.
Having a \textit{social platform digital twin} would allow researchers to perform different social experiments, investigate feedback loops between social and technical components of such systems, test different what-if scenarios, use insights to inform policy makers, in a more cost-effective and ethically-compliant way. 
However, for the additional complexity of the system \textit{society}, creating such a twin is not a trivial task. 
Efforts have been made in this direction by defining a digital twin for complex networks systems, where data from the real systems are used to represent and model the digital system, mimicking both network characteristics and processes (e.g. spreading) \cite{wen2023digital}. 
Simulations in \cite{wen2023digital} involve both network evolution 
and infection spreading processes to assess network resilience. 
Other studies move towards the creation of a human digital twin, a digital replica of a real human. 
Besides replicating physical characteristics, as in healthcare, replicating cognitive capabilities and heterogeneous behaviours is the greatest challenge of creating a comprehensive twin of online social platforms. 
LLM-enhanced agents have been extensively explored 
in their ability to autonomously replicate human characteristics when prompted and/or fine-tuned. 
However, to date there has been no \textit{social platform digital twin} mirroring characteristics of one or more platforms, and enriched with agents that mimic human behaviour - which is the core of our proposal. 

\newpage
\section{Y Social - Digital Twin}
\label{sec:Y}

{\tt Y Social} is meant to allow for profound flexibility in simulation scenario development.

To fulfill such a goal, we designed our social media platform digital twin to be composed of three interacting modules: 
\begin{itemize}
    \item[i.] a REST API server designed to expose all those primitives describing the actions implemented by the social platform and to store the simulation data;
    \item[ii.] a Large Language Model (LLM) server that serves requests related to agent interrogations (e.g., simulating decision-making protocols, text generations\dots); 
    \item[ii.] a simulation client that implements the agent logic and acts as a middle layer that interfaces the REST API server with the LLM (Large Language Model) one.
\end{itemize} 

\begin{figure}[t]
    \centering    \includegraphics[width=0.9\linewidth]{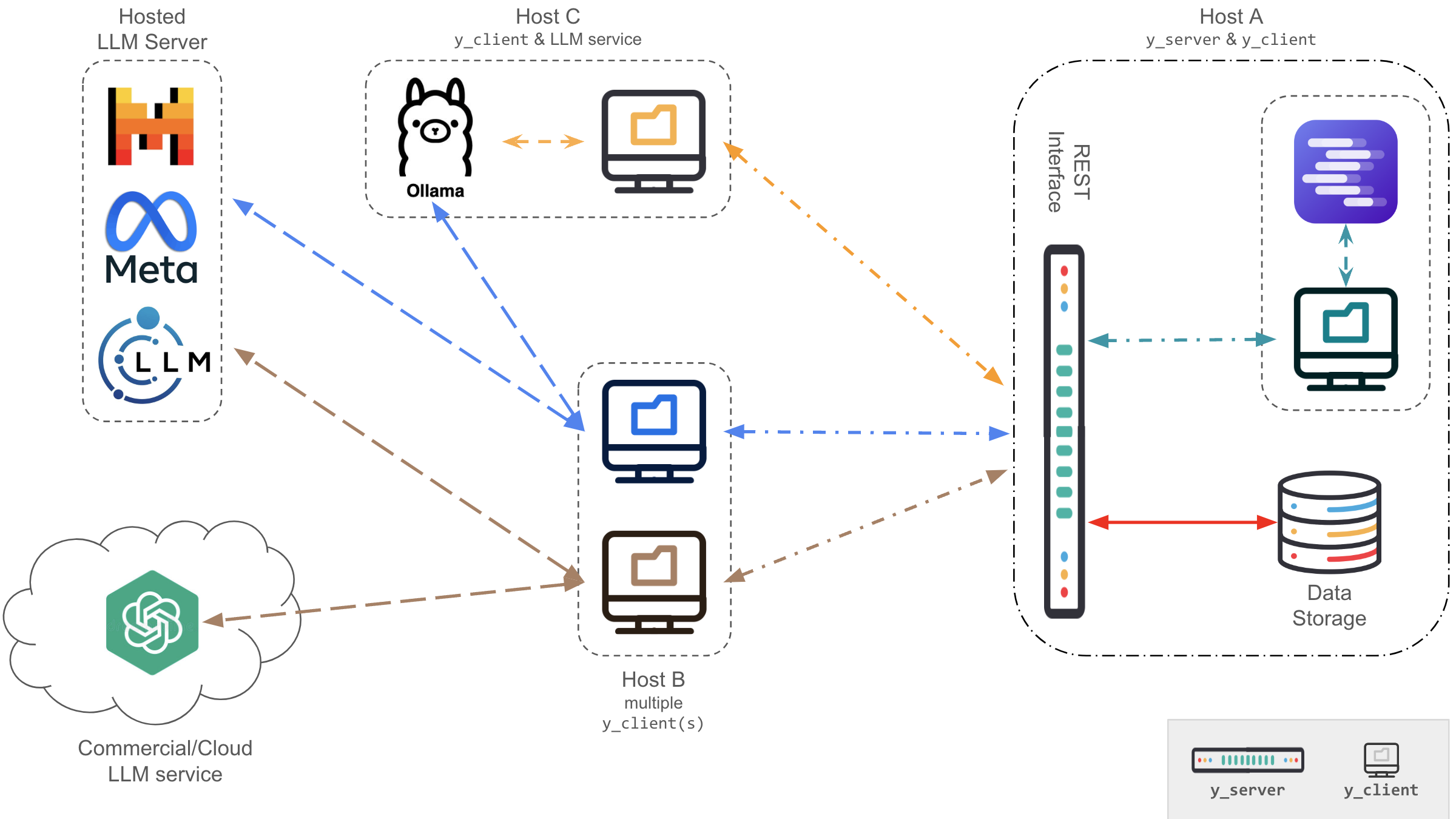}
    \caption{{\bf Y architecture.} The Digital Twin is composed of a {\tt y\_server} - exposing a REST API allowing simulation synchronization and data storage - and a {\tt y\_client} implementing the simulation logic and the interface toward the {\tt LLM service(s)} simulating agents' behaviors. The {\tt y\_client} can be deployed: on the same machine of the {\tt y\_server} (Host A); on a machine running multiple clients (Host B); on a different machine hosting also the {\tt LLM service} (Host C). Moreover, by design, a generic {\tt y\_client} can leverage one (or more) LLM(s), either self-hosted or commercial.}
    \label{img:structure}
\end{figure}

Given its modular design, as visually simplified in Figure \ref{img:structure}, {\tt Y} allows several clients to coordinate during every simulation, thus distributing the computational power needed to simulate agents' behaviors.
It is worth underlining that {\tt Y} makes it possible to leverage commercial models (i.e., OpenAI) as well as self-hosted ones\footnote{Under the assumption that the LLM server is OpenAI API compliant.} (e.g., served through ollama, LM Studio or similar services) as a LLM server.
Moreover, if specified in the simulation configuration, each {\tt Y} client can assign multiple LLM models to the simulated agents, thus generating a population reflecting heterogeneous behaviors, linguistic capabilities, and adherence to the provided \emph{profiles}.

In the following, we discuss the REST API server (i.e., {\tt y\_server}) and the simulation client (i.e., {\tt y\_client}) modules, describing their rationale and the implementation choices made while implementing them.

\subsection{{\tt y\_server}: Social Media platform primitives}

To properly describe a Social Media digital twin, the first thing to specify is the primitives that the agents can use to describe their social actions.
We designed Y's primitives to resemble the ones offered by platforms like X/Twitter, Mastodon, and BlueSky Social.
In particular, we defined the following REST endpoints to identify agents actions:

\begin{itemize}
    \item[] {\tt /read}: returns a selection of agent posts as filtered by a specified content recommender system;
    \item[] {\tt /post}: registers a new post on the database (along with all the metadata attached to it - e.g., mentions, hashtags, inferred emotions\dots);
    \item[] {\tt /comment}: makes its possible to register a comment to existing user-generated content (along with its metadata);
    \item[] {\tt /reply}: provides a (recommender system-curated) list of posts that mention a given agent;
    \item[] {\tt /news}: allows agents to publish news gathered from an online resource (e.g., from a news media outlet), adding a personal comment to it;
    \item[] {\tt /share}: allow agents to share previously published news articles generating new discussion threads;
    \item[] {\tt /reaction}: registers the reactions (e.g., like/dislike) of a given agent to a given content on the database;
    \item[] {\tt /follow\_suggestions}: provides a selection of potential contacts for a target agent leveraging a recommender system;
    \item[] {\tt /follow}: allows agents to establish new social connections (follow) or break existing ones (unfollow).  
\end{itemize}

{\tt y\_server} exposes additional endpoints to handle system authentication, other articulated actions (e.g., timeline construction, content/follow recommendation pipelines), and multiple client synchronization.
We now focus on {\tt Y}'s algorithmic curation from the perspective of the platform-user dynamics, namely content and follower recommendation pipelines.

\subsubsection{Introducing Algorithmic Bias: Recommender System(s)}

\noindent{\bf Content Recommendations.} Several of the actions introduced - namely, {\tt read}, {\tt comment}, {\tt reaction}, {\tt share}, {\tt reply} - focus on allowing agents to ``react" to contents produced by peers.
Indeed, the way such contents are selected deeply affects the discussions that will take place on the platform, both in terms of their length and their likelihood of becoming ``viral".
For such a reason, {\tt Y} natively integrates several standard recommender systems for content suggestion (and allows for an easy implementation of alternative ones), namely:
\begin{itemize}
    \item {\tt Random}: suggests a random sample of $k$ recent agents' generated contents;
    \item {\tt ReverseChrono}: suggests $k$ agents' generated contents in reverse chronological order (i.e., from the most recent to the least recent);
    \item {\tt ReverseChronoPopularity}: suggests $k$ recent agents' generated contents ordered by their popularity score computed on as sum of the like/dislike received;
    \item {\tt ReverseChronoFollowers}: suggests recent contents generated by the agent's follower - it allows to specify the percentage of the $k$ contents to be sampled from non-followers;
    \item {\tt ReverseChronoFollowersPopularity} suggests recent contents generated by the agent's follower ordered by their popularity - it allows to specify the percentage of the $k$ contents to be sampled from non-followers;
\end{itemize}
Each content recommender system is parametric on the number $k$ of elements to suggest.
To increase the scenario development potential of {\tt Y} (e.g., to design A/B tests), each instance of the simulation client ({\tt y\_client}) can assign a specific instance/configuration of the available recommender systems to each of the generated agents.
\\ \ \\
\noindent{\bf Follows Recommendations.} Among the described agent actions, a particular discussion needs to be raised for the {\tt follow} one.
{\tt Y} agents are allowed to establish (and break) social ties following two different criteria: i) as a result of a content interaction (e.g., after the evaluation of a content posted by a peer); ii) selecting a peer to connect with among a shortlist proposed by a dedicated recommender system.

As for the content recommendations, {\tt Y} integrates multiple strategies to select and shortlist candidates when an agent $a$ starts a {\tt follow} action.
\begin{itemize}
    \item {\tt Random}: it suggest a random selection of $k$ agents;
    \item {\tt Common Neighbours}: it suggests the top $k$ agents ranked by the number of shared social contacts with the target agent $a$ \cite{newman2001clustering};
    \item {\tt Jaccard}: it suggests the top $k$ agents ranked by the ratio of shared social contacts among the candidate and the target agents over the total friends of the two \cite{chowdhury2010introduction};
    \item {\tt Adamic Adar}: the top $k$ agents are ranked based on the concept that common elements with very large neighborhoods are less significant when predicting a connection between two agents compared with elements shared between a small number of agents \cite{adamic2003friends};
    \item {\tt Preferential Attachment}: it suggests the top $k$ nodes ranked by maximizing the product of $a$'s neighbor set cardinality with their own \cite{newman2003structure}.
\end{itemize}

Each of the implemented methodologies, borrowed from classic unsupervised link prediction scores \cite{liben2003link}, allow agents to grow their local neighborhood following different local strategies - each having an impact on the overall social topology of the system (e.g., producing a heavy-tailed degree distribution \cite{barabasi2003scale, barabasi2009scale}).
Moreover, {\tt Y} allows for specifying if the follower recommendations should be biased (and to what extent) toward agents sharing the same political leaning so as to implement homophilic connectivity behaviors.

\begin{figure}
    \centering
    \includegraphics[width=0.9\linewidth]{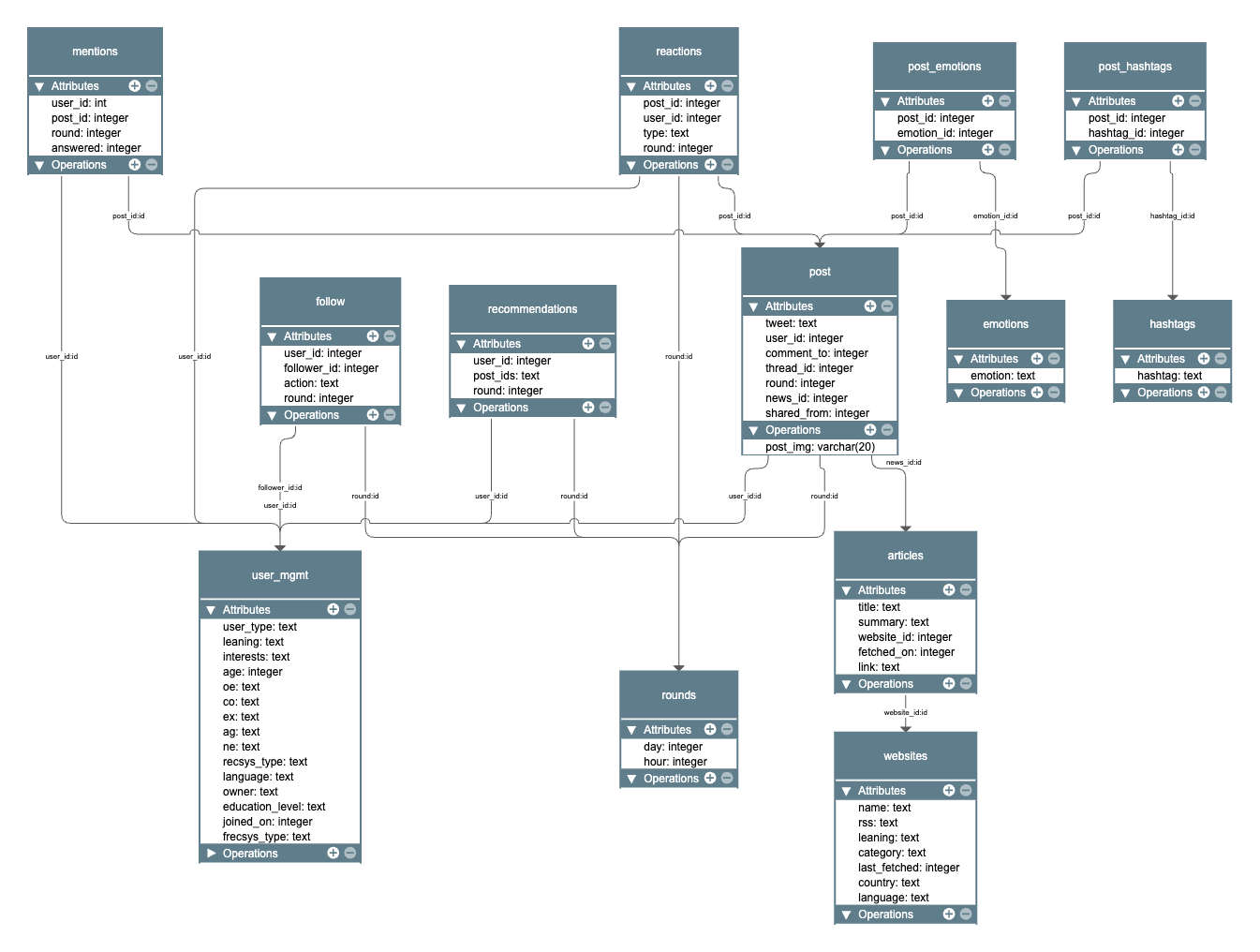}
    \caption{{\bf Y database diagram}. The Digital Twin simulation data is stored in a database where the main tables are: {\tt user\_mgmt} describing the agents' profiles; {\tt post} collecting the data of the agents' generated contents (often characterised by many-to-many relations - e.g., toward the {\tt mentions}, {\tt hashtags}, {\tt reactions} and {\tt emotions} tables); {\tt websites} and {\tt articles} storing the updated information on the shared news; {\tt follow} storing the established/removed social relationships; {\tt rounds} simulating the system the temporal clock - used to enforce {\tt y\_client(s)} synchronization.}
    \label{img:db}
\end{figure}

\newpage
\subsection{{\tt y\_client}: Simulating Agents' social interactions}
\label{sec:agents}
As discussed, the primary function of {\tt y\_server} is to expose, through a REST API, access to all those primitives that allow for the execution of the Digital Twin simulation.
If {\tt y\_server} can be seen as a ``light" interface toward the database storing the simulation results (see Figure \ref{img:db} for an overview of the DB diagram), the {\tt y\_client} is the module that implements most of the business logic of the Y.

In particular, {\tt y\_client} defines the simulation agents and their characteristics, implements the agents' actions selection process (i.e., the strategy used to select which endpoint exposed by {\tt y\_server} to connect to and how to interpret the obtained responses), handle the access to external online resources and, finally orchestrate the simulation by synchronizing with other clients' activities through dedicated {\tt y\_server}'s primitives.
In the following, we review the main facilities offered for each task - starting from the most important component of a social media platform Digital Twin: the agents that populate it.
\subsubsection{LLM-powered agents}
\label{sub:agents}
Y's main peculiarity lies in the kind of agents that independently perform the actions it exposes.
Conversely, from a classic mechanistic simulation environment, where action selection is often implemented as a sampling from a given probability distribution - that implicitly fixes their expected volume distribution - {\tt Y} leverages LLMs as \emph{chaos engines}, leaving each agent the \emph{choice} w.r.t. the actions to take.

Moreover, to increase the heterogeneity of agents' behaviors - and generated contents -, {\tt Y} leverages the ability of LLMs to \emph{impersonate} different users' profiles (\emph{persona}).

Through agent profiling, {\tt Y} allows for control of the population segmentation over multiple dimensions (that can be expanded if needed by the specific simulation scenario).
\\ \ \\
\noindent{\bf Agent profile.}
While instantiating an agent, {\tt y\_client} allows to specify the following descriptive dimensions:
\begin{itemize}
    \item[-] {\tt Agent Name and Ownership}: Name is the unique identifier for the generated agent, while ownership links the agent to the client execution that generated it. This latter information is useful for checking the simulation status while multiple clients (having different configurations) are active.
    \item[-] {\tt LLM model}: This field identifies the model used for interpreting the agent's actions. Each agent can be impersonated by a different model (e.g., mistral, llama3, phi3, chatgpt\dots).
    \item[-] {\tt Age, Spoken Language(s), Education level}: Such demographic information has been shown to affect the content generated by LLMs. In particular, Age and Education can be used to tune the expected writing styles of agents. At the same time, the list of spoken languages will filter the contents and (the modality) the agents interact with.
    \item[-] {\tt Political leaning}: This dimension is crucial for social simulations to generate social interactions centered on socio-political themes. It can be used to coarse-grain pre-set the expected agent's opinion.
    \item[-] {\tt Topic(s) of Interest}: The list of interest topics represents the background of the agent and acts as a starting point for the generation of posts and to make consistent contributions to discussion threads. Moreover, specifying a reasonable set of topics per agent increases the heterogeneity of produced texts.
    The global topic selection also allows for the setup of topical-oriented simulations (e.g., describing a digital twin of a political forum rather than of a self-help community).
    \item[-] {\tt Recommender system(s)}: As previously described, each agent is allowed to decide - at instantiation time - which content/follow recommendation strategies to leverage among the ones provided by the {\tt y\_server}.
    \item[-] {\tt Big Five personality traits}: The Big Five personality traits, also known as the Five-Factor Model (FFM \cite{mccrae1992introduction}), describe five broad dimensions of human personality. Such five dimensions are used to fine-tune the agent's expected behavior while interacting with other users' contents (and generating its own).
\end{itemize}

Focusing on the last dimension, each agent is characterized by five dichotomic variables, each capturing the extreme values of one of the Big Five personality traits, namely:

\begin{itemize}
    \item[-] Openness to Experience (oe): This trait features characteristics such as imagination, insight, and a broad range of interests. Highly open people are often creative, curious, and open to new experiences. Those low in this trait may be more conventional and prefer routine.
    \item[-] Conscientiousness (co): This dimension involves high levels of thoughtfulness, good impulse control, and goal-directed behaviors. Highly conscientious people tend to be organized, mindful of details, and responsible. Those low in Conscientiousness may be more spontaneous and less structured.
    \item[-] Extraversion (ex): This trait includes excitability, sociability, talkativeness, assertiveness, and high emotional expressiveness. Extroverted individuals are energetic and enjoy being around people. Introverts, who are lower in extraversion, may be more reserved and prefer solitary activities.
    \item[-] Agreeableness (ag): This personality dimension includes trust, altruism, kindness, and affection. Agreeable individuals tend to be cooperative, compassionate, and good-natured. Those lower in Agreeableness may be more competitive and sometimes less empathetic.
    \item[-] Neuroticism (ne): This trait refers to the tendency to experience negative emotions, such as anger, anxiety, or depression. Individuals high in Neuroticism are more likely to experience emotional instability and mood swings. Those low in this trait are generally more stable and resilient.
\end{itemize}

These five traits are considered relatively stable and provide a comprehensive framework for understanding human personality.
{\tt Y} allows each agent to be characterized by a high/low value for each of the five traits, allowing for 32 different personalities.

The defined agent profile dimensions concur with the description of detailed and consistent characters that the LLM can interpret in its role-playing.
In particular, defining the Big Five personality trait profiles affects agents' choices in content generation modality.

In order to leverage the defined profile, each LLM action prompt is anticipated by the following role-play directives:

\begin{tcolorbox}
\begin{verbatim}
You are a {age} year old {leaning} interested in {",".join(interest)}. 
Your Big Five personality traits are: {oe}, {co}, {ex}, {ag} and {ne}.
Your education level is {education_level}.
            
Act as requested by the Handler. 
- DO NOT refuse to generate a response. 
- DO NOT generate unacceptable content but act coherently with your character profile.
- DO NOT describe your profile in the generated texts.
- All generated texts MUST be short (up to 200 characters).
\end{verbatim}
\end{tcolorbox}
\noindent{\bf Actions and Prompts.} 
Once the agents' traits are specified, we can move on to discuss how the {\tt y\_client} implements the actions associated with the primitives offered by the {\tt y\_server} and, therefore, how each agent behaves during a generic simulation round.

The entry point during each agent simulation round is the {\tt select action} procedure. 
In that step, the LLM interpreting the agent is prompted with a request to select an action from a predefined (and customizable) list.

\begin{tcolorbox}
\begin{verbatim}
Select a word randomly from the following list and write it. 
Do not write additional text.

## START INPUT
{actions}
## END INPUT
\end{verbatim}
\end{tcolorbox}

Once the action(s) are selected - depending on the configuration, each agent can perform one or multiple actions per round - the {\tt y\_client} routine associated with such action is activated.
In the following, we report the details of each action implementation and - where present - the LLM prompt assigned to it.
\\ \ \\
\noindent{\tt READ.} The read action allows agents to \emph{react} to peers' posts/comments.
Firstly, a shortlist of agents' posts/comments is computed leveraging the content recommender system associated with the agent. An element from such a list is randomly chosen (to avoid a self-reinforcing selection bias favoring a single element), and the following prompt is activated:

\begin{tcolorbox}
\begin{verbatim}
Read the following text, write YES if you like it, NO if you don't, NEUTRAL otherwise.

## START TEXT
{post_text}
## END TEXT
\end{verbatim}
\end{tcolorbox}
The outcome of the prompt is then used to associate a like/dislike reaction to the selected content from the active agent.
Multiple reaction types can be easily integrated by adapting the prompt request.

Additionally, depending on the reaction outcome, the LLM agent is prompted with a similar request asking about the intention of following/unfollowing the author of the post/comment.
Such additional steps ensure the dynamicity of the social interactions, allowing them to tie them to explicit topology-driven suggestions of the following recommender system and the contents accessed and their evaluation.
\\ \ \\
{\tt POST.} This action allows agents to start a new discussion focused on a randomly selected sample of their interest topics (sample specified in the role-play pre-prompt).

\begin{tcolorbox}
\begin{verbatim}
Write a short tweet introducing a topic of interest to you. 

- Be consistent with your Big Five personality traits.
- Avoid excessive politeness.
- Do not exceed the limit. Make it short.
- Write in {language}.
\end{verbatim}
\end{tcolorbox}

The term \emph{tweet} is used in place of the more generic \emph{text}/\emph{post} to implicitly allow the LLM to recover information on the expected content structure.

Once generated, the post elements such as mentions (identified by the starting {\tt @} character) and hashtags {identified by the starting {\tt \#} character) are extracted.
Moreover, after the text generation phase, the LLM is also asked to annotate the produced text with the emotions it elicits - as categorized by the GoEmotions taxonomy\cite{demszky2020goemotions}.
The same post-processing step is applied to all generated texts.
\\ \ \\
{\tt COMMENT/REPLY/SEARCH.}
Similarly to the {\tt READ} action, the content to be commented is identified by the selected recommender system.
After selecting, the following prompt (along with the post-processing already discussed for the {\tt POST} action) is applied.

\begin{tcolorbox}
\begin{verbatim}
Read the following conversation and add your contribution to it.

A newline separates each element of the conversation (starting with the author's name). 
            
- You can tag the author of the tweet using @.
- Be consistent with your Big Five personality traits.
- Avoid excessive politeness.
- Your comment MUST contribute to the conversation.
- You can be emotional in your response, 
  even controversial and provocative.
- You are a native speaker of the {language} language: 
  if the original post is not written in {language}, answer 
  assuming a non-native proficiency.

            
## START CONVERSATION
{conversation}
## END CONVERSATION
\end{verbatim}
\end{tcolorbox}

As reported in the prompt, the LLM agent is fed not only with the selected content but also with a limited view of the posts that precede it in the conversation thread (with a window parameterized during the configuration phase).
Moreover, the prompt specifies how to handle contents that are not written in the agent-selected language(s).
Specifying the author's names (i.e., their handles) in the input conversation allows consistent user tagging.

Conversely, from the {\tt COMMENT} action, {\tt REPLY} constrains the recommender system to select a conversation where the agent was (recently) mentioned.
Such additional action allows consistency and articulation of discussion threads.
Similarly, {\tt SEARCH} allows the agent to discover new content/peers to interact with based on its interests.
Instead of constraining the recommender system by the received mentions, in {\tt SEARCH}, the filter focuses on content sharing (some of) the hashtags that the agent used to annotate its recent posts.
\\ \ \\
{\tt FOLLOW.} This action, not tied to the outcome of an LLM prompt, implements the creation of social ties after a recommender system suggestion.
Suggestions are ranked by the scoring function implemented by the agent following the recommender system, then the scores are transformed into probabilities, and a biased random selection is performed to select the peer to connect to.
It is worth noticing that such an action is not tied to the content shared by the shortlisted agents but rather to the social system topology.
Moreover, depending on the content recommender system, expanding the agent's social neighborhood might impact the contents he will potentially be exposed to.
\\ \ \\
\noindent{\bf Online News.} The last two actions focus on a specific functionality {\tt Y} offers: allowing LLM agents to access, comment, and share news gathered from selected websites (e.g., online news sources).
Such functionality has been integrated to allow synthetic agents not only to linchpin discussions on their favored topics but also to instantiate conversations centered on external inputs.

News access is obtained through an RSS (Really Simple Syndication) feed parser module. 
During the simulation configuration phase, it is possible to specify the RSS feeds available to the LLM agents ({\tt Y} comes with more than 600 validated RSS feeds focused on international politics and science/pseudoscience communication - each annotated with the political leaning of the news media).
Leveraging such information, an agent can: 
\begin{itemize}
    \item[i.] collect the daily news posted by an online media and publish a post commenting on one of them (through the {\tt NEWS} action);
    \item[ii.] {\tt SHARE}, in a new thread, a piece of news previously posted by a peer.
\end{itemize}
As discussed, the first action allows the integration of real news into the simulated debate. 
Moreover, it is possible to specify the preferred category (e.g., politics, sports, science) and/or political leaning of the news media to be accessed by each agent to generate specific posting types (i.e., having agents that broadcast and support news aligned with their political views or others that share a piece of news from an ideologically distant outlet to criticize it).
The following prompt characterizes a general {\tt NEWS} action:

\begin{tcolorbox}
\begin{verbatim}
Read the title and summary of the following article and share your thoughts about it. 

- Be consistent with your Big Five personality traits.
- Avoid excessive politeness.
- Do not exceed the limit. Make it short.
            
## START INPUT
Title: {article.title}
Summary: {article.summary}
## END INPUT
\end{verbatim}
\end{tcolorbox}

The second action allows the generation of content diffusion processes within the simulated social space.
Each time an agent {\tt SHARE} a previously posted content, it adds its comment on the news (and to the original poster comment).
Even if the shared content initiates a novel discussion thread, {\tt y\_server} traces the relationship between the original and the shared contents.

\newpage
\subsubsection{Orchestrating a Simulation.}
Once the rationale behind the agents' characterization and the rules that govern their behavior is defined, we discuss how (each instance of) {\tt y\_client} orchestrates the agents' activities.
To such an extent, we review the configuration variables exposed by {\tt y\_client} and briefly discuss how the simulation loop unfolds.
\\ \ \\
\noindent{\bf Simulation Configuration.}
{\tt y\_client} configuration file (i.e., the simulation \emph{recipe}) allows specifying parameters for four entity types: servers, simulation, agents, and posts.
\begin{tcolorbox}
\small
\begin{verbatim}
{
  "servers": {
    "llm": "http://127.0.0.1:11434/v1",
    "llm_api_key": "NULL",
    "api": "http://127.0.0.1:5000/"
  },
  "simulation": {
    "name": "experiment_name",
    "client": "YClientBase",
    "days": 365,
    "slots": 24,
    "starting_agents": 1000,
    "new_agents_per_iteration": 10
    "hourly_activity": {...},
  },
  "agents": {
    "education_levels": ["high school", "bachelor", ...],
    "languages": ["english"],
    "max_length_thread_reading": 5,
    "reading_from_follower_ratio": 0.6,
    "political_leanings": ["Democrat", "Republican", ...],
    "age": {"min": 18, "max": 80},
    "round_actions": {"min": 1, "max": 3},
    "nationalities": ["American", "Italian", ...],
    "probability_of_daily_follow": 0.1,
    "llm_agents": ["llama3", "mistral", ...],
    "n_interests": {"min": 4, "max": 10},
    "interests": [...],
    "big_five": {...}
  },
  "posts": {
    "visibility_rounds": 36,
    "emotions": {...}
  }
}        
\end{verbatim}
\end{tcolorbox}
Server configuration allows specifying the addresses (and ports) for both the REST API service (namely, {\tt y\_server}) and the LLM service - along with the API key in case of commercial services.
Simulation configuration defines the simulation duration, number of starting agents, and novel agents spawn per day. 
In particular, the shown configuration file specifies that the simulation will cover 100 days, each composed of 24 iterations (one for each hour), for a total of 2400 \emph{rounds}.
Moreover, the {\tt hourly\_activity} parameter makes it possible to specify the percentage of agents to be considered active during each daily slot; simulation {\tt name} assigns an identifier to the scenario (allowing multiple clients to coordinate on a same simulation); {\tt client} specifies the implementation to be used (opening for future extensions of {\tt y\_client} functionalities).
Concerning the {\tt hourly\_activity} parameter, {\tt Y} is pre-set with an expected hourly activity rate fitted on a yearly dataset covering 80\% of Bluesky Social users population activities \cite{failla2024m}.

The subsequent configuration section describes the parameter ranges that characterize the LLM-powered agents.
Although each agent profile can be handcrafted in all its peculiarities by the user, {\tt y\_client} allows automatically generating a population of agents by sampling their features from the provided values described in the configuration file.
Apart from the agent's characteristics already described in Section \ref{sub:agents}, the configuration file allows for the specification of the maximum width of thread posts to read before commenting ({\tt max\_length\_thread\_reading}) and (if needed) the percentage of posts from the followee that the content recommender system can suggest ({\tt reading\_from\_followee\_ratio}).
Finally, the post configuration section allows specifying the visibility in terms of rounds of published content (e.g., in the provided example, 36 rounds, or one day and a half) along with other additional information like the emotion dictionary for post-processing annotations.
\\ \ \\
\noindent{\bf Simulation workflow.}
Algorithm \ref{alg:loop} reports a prototypical simulation workflow implemented using the API offered by {\tt y\_client}. 
The simulation starts by generating the agents and registering them to the {\tt y\_server} (in case of resumed simulation, pre-existing agents are passed as inputs and verified on the server. 
For the sake of simplicity, such an alternative path is not reported in the pseudocode).
After the initialization phase, the simulation loop unfolds by iterating over the selected period and activating a percentage of agents coherent with what is specified in the configuration file for each slot.
Before each round activity, the {\tt y\_client} instance synchronizes with the {\tt y\_server} simulation clock to avoid issues in distributed client settings.
At the end of each day, the agent population is increased accordingly to what is specified by the {\tt new\_agents\_per\_iteration} simulation parameter.

It is worth noticing that, in addition to the configuration described in the previous section, it is possible (by slightly modifying the inner loop code) to tune the actions available to the agents and to specify the number of actions per round.
In the current example, an active agent can execute up to three different actions per round.

\begin{algorithm}[t]
\DontPrintSemicolon
  
  \KwInput{{\tt config}: Simulation configuration Files}
  \KwInput{{\tt feeds}: RSS feeds}

  \tcc{configuring agents and servers}
  agents $\leftarrow$ create\_agents({\tt config}, {\tt feeds})\\
  y\_server $\leftarrow$ connect({\tt config}.servers.api)\\

  \tcc{simulation loop}
  \For{day $\in$ range({\tt config}.simulation.days)}{
    \For{slot $\in$ range({\tt config}.simulation.slots)}{
        \tcc{synchronize with the {\tt y\_server} clock}
        h = {\tt y\_server}.get\_current\_slot()\\
        \tcc{identify the active agents for the current slot}
        expected\_active = int(len(agents) * {\tt config}.simulation.hourly\_activity[h])\\
        active = random.sample(agents, expected\_active)\\

        \For{agent $\in$ active}{ 
        \tcc{evaluate agent's actions (fitted on the agent's rounds\_actions value)}
            agent.select\_action(["NEWS", "POST","COMMENT",\\ 
            "REPLY", "SHARE", "READ",\\ "SEARCH", "NONE"])
        }
    }
    \For{agent $\in$ agents}{ 
    \tcc{evaluate following (only for probability\_of\_daily\_follow among the active agents)}
        agent.select\_action(["FOLLOW", "NONE"]) 
    }
    \tcc{increase the agent population (if specified in {\tt config})}
    agents.add\_new\_agents()
}
\caption{{\tt y\_client} default implementation}
\label{alg:loop}
\end{algorithm}
\newpage
\section{Case study: Political Debate Arena}
\label{sec:case}
To briefly showcase the data a Y simulation makes available, in this section, we describe a simple case study involving a relatively small number of LLM agents debating general politics-related themes.
To such an extent, we first describe the simulation configuration, \ref{sec:toy_parameters}, then produce a simple analysis of the generated outputs \ref{sec:toy_analysis}.

\subsection{Simulation configuration}
\label{sec:toy_parameters}
The performed simulation involves an initial population of 1000 interacting LLM agents that - during 100 days (i.e., 2400 hourly rounds) - grew to 2000 (i.e., introducing ten new agents daily).
For simplicity's sake, all agents were powered by the same LLM model, ({\tt llama3-7b} \cite{llama3modelcard} served using ollama\footnote{\url{https://ollama.com/}} on consumer-level hardware), and configured to leverage the same content  ({\tt ReverseChronoFollowersPopularity}) and follower ({\tt PreferentialAttachment}) recommender system.
Moreover, the latter recommender was parametrized to guarantee that up to 60\% of the recommendations come from content generated by already followed agents.

To ensure that the discourse focused on politics-related debates, we imposed agent interests sampled from a predefined list of discussion arguments (e.g., education, minority discrimination, economics, healthcare, welfare, justice\dots). 
Moreover, we randomly assigned each agent to a political leaning among the following classes: Democrat, Republican, Libertarian, Green, Independent, Alt-Right, Alt-Left, Anarchist, and Centrist.
Other agents' characteristics (e.g., age, education level, Big Five personality traits) were randomly sampled from predefined ranges/value lists.
Agent profiles were automatically generated leveraging the {\tt faker}\footnote{\url{https://faker.readthedocs.io}} Python library.

Agent-generated content was made available as recommendation candidates for 36 slots (one day and a half) after their original publishing. 
The access to external news sources was implemented by a curated list of 600 RSS feeds targeting major political news outlets, known conspiracy/fake news websites, and science-related online sources (all of them annotated - whenever possible - with their political alignment as reported by Media Bias Fact Check\footnote{\url{https://mediabiasfactcheck.com/left/}}. 
Generated data and simulation configuration files are available on the {\tt Y Social} website\footnote{Political debate arena data: \url{https:/YSocialTwin.github.io}}.

\subsection{Generated Data Examples}
\label{sec:toy_analysis}
Figure \ref{img:rates} reports a descriptive statistics sub-sample of the agent population. 
Figure \ref{img:rates}(a) shows the LLM agents' hourly activity rates as fitted by a data-driven observation made on the full dump of a similar service, BlueSky \cite{failla2024m}.
That choice was made to better simulate agents' activities' circadian cycles, as well as realistically modulate the expected number of online users. 
Figure \ref{img:rates}(b-c) reported the political leaning and age distribution: in line with the purpose of the case study - developed to showcase (part of) the data {\tt Y} can generate, not to address specific research questions - neither distribution was skewed toward specific leanings/age groups. 

Let us now focus on the contents produced by {\tt Y} agents during the simulation.
Figure \ref{img:data}(a) reports the cumulative decreasing distribution of agents' published contents, breaking it down into the following categories: Posts, Comments, News Articles, Shared Articles, hashtags used, and Mentions made and received.
As can be observed - and aligned with what is expected in online social media platforms - agents tend to post more than they comment, with 80\% of them individually generating more than 30 posts, against a volume of only 15 comments each.
Similar trends follow the News Articles (i.e., those posts sharing news for the first time) and the Shared Articles (i.e., those posts reposting news articles posted by other agents).

\begin{figure*}
    \centering
    \subfloat[]{\includegraphics[width=.97\linewidth]{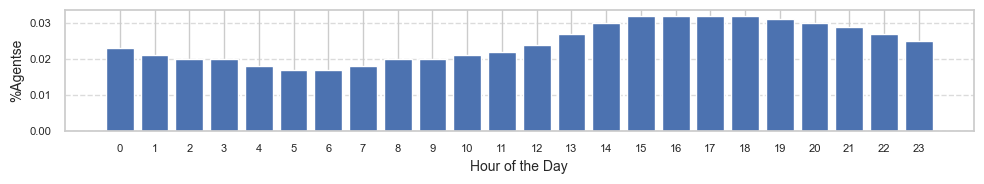}}\\
    \subfloat[]{\includegraphics[width=0.45\linewidth]{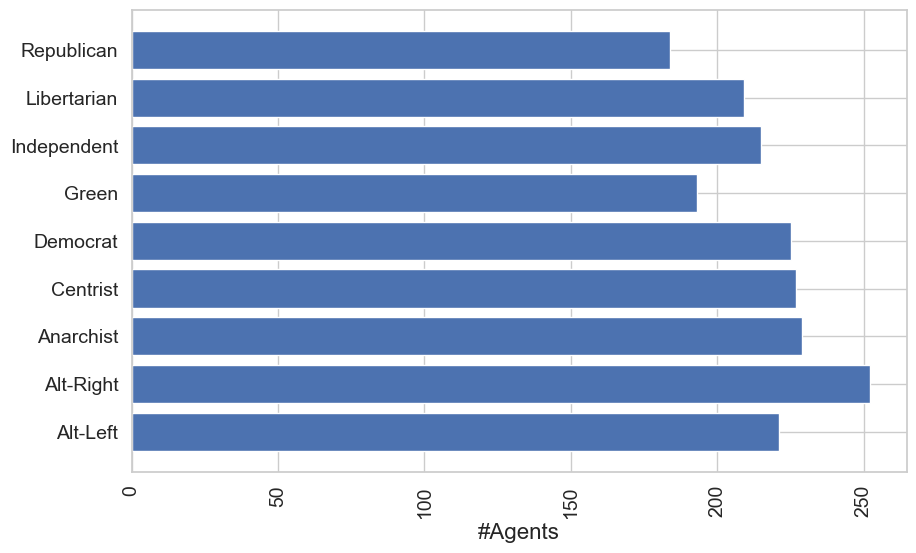}} \qquad \quad
    \subfloat[]{\includegraphics[width=0.45\linewidth]{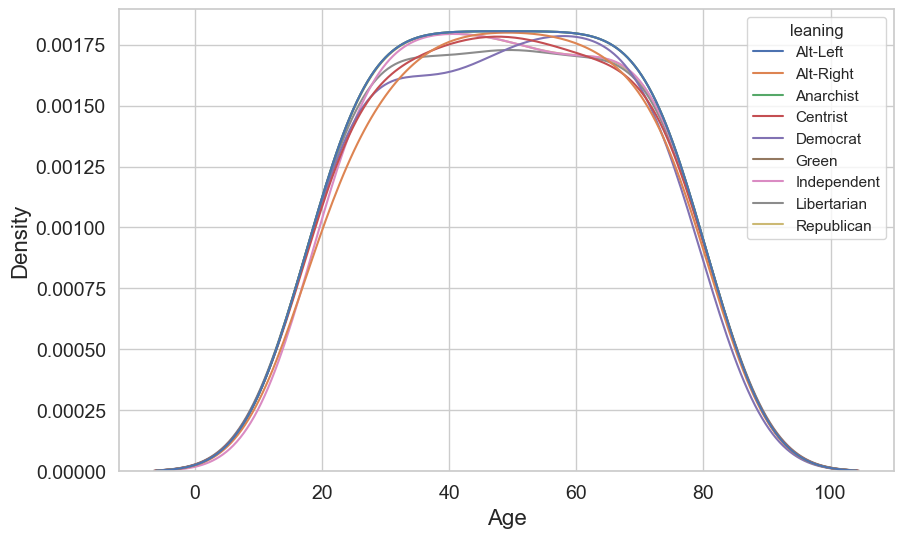}}
    \caption{{\bf Simulation setup.} (a) Hourly activity rates of LLM agents (fitted on BlueSky Social data); (b) Representativeness of political leaning classes in the simulated agent population; (d) Age distribution per political leaning.}
    \label{img:rates}
\end{figure*}

\begin{figure*}[]
    \centering
    \subfloat[]{\includegraphics[width=0.46\linewidth]{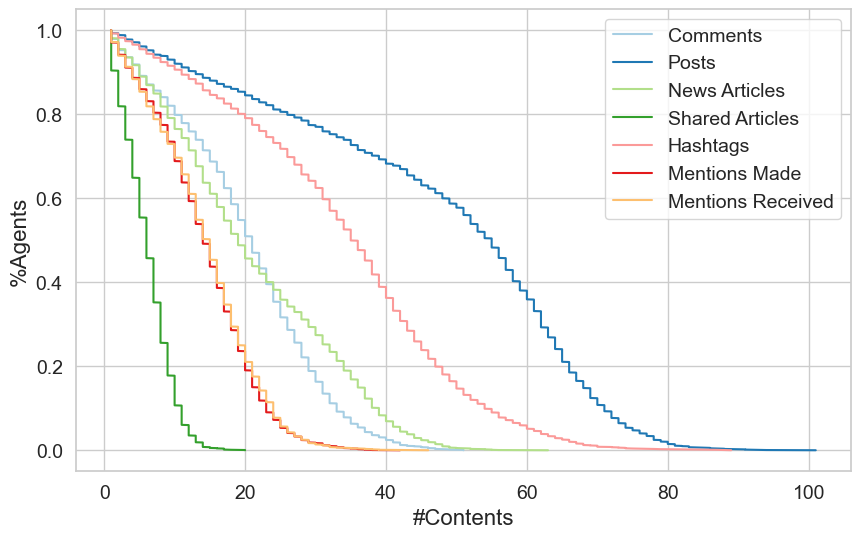}} \qquad\quad
    \subfloat[]{\includegraphics[width=0.47\linewidth]{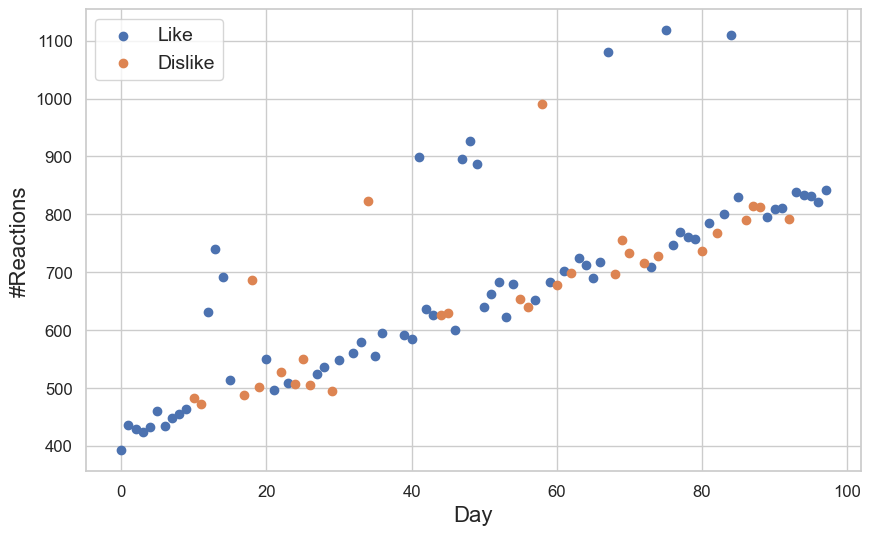}}
    \\
    \subfloat[]{\includegraphics[width=0.50\linewidth]{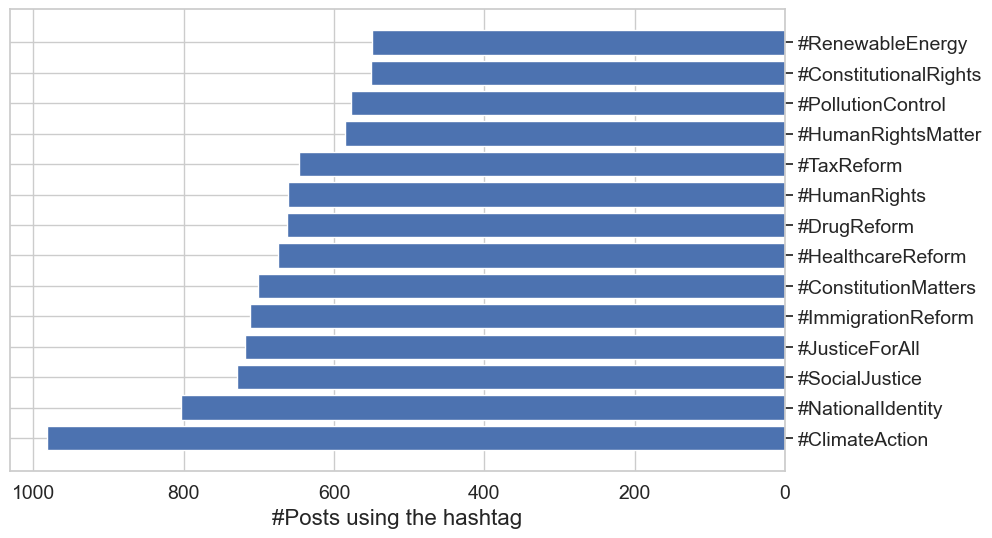}}
    \subfloat[]{\includegraphics[width=0.48\linewidth]{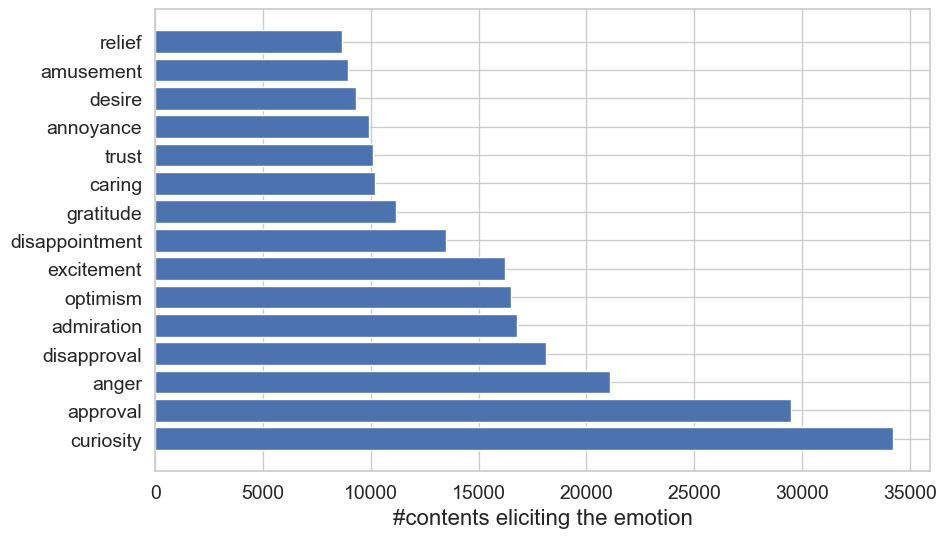}}   
    \caption{{\bf Generated content statistics.} (a) CCDF of generated contents (posts, comments, news, hashtags, mentions) per agent; (b) temporal distribution of agents' reactions (like/dislike) to peer contents; (c) Most used hashtags; (d) Most frequently elicited emotions in generated texts.}
    \label{img:data}
\end{figure*}

With regards to the mention (i.e., tagging peers within a comment to specify its target), it seems that {\tt Y} agents tend to identify small and closed-knit circles of peers to interact with - as shown by the cumulative decreasing distribution of the unique usernames mentioned per agent.
{\tt Y} agents extensively used hashtags to annotate their posts/comments with topical keywords (with 60\% of agents using more than 30 unique hashtags each).

Figure \ref{img:data}(c) details the most used hashtags' variety, volume and simulation-topic relatedness.
Agents' debates focused on themes like Climate,  Social Justice, Healthcare, Immigration, Human Rights, Pollution control and Renewable energy - all topics that perfectly align with policymaking and political issues.
Moreover, Figure \ref{img:data}(b) underlines the trends of agents' reactions to published contents in time.
The daily increasing trends for both likes and dislikes are expected due to the incrementally growing number of agents during the simulation: peaks are due to viral content, thus experiencing a visibility boost due to the selected recommender algorithm.
Moreover, it is worth noticing how agents provide a reasonable volume of negative feedback (i.e., dislikes) - something not trivial to observe due to LLMs' tendency toward politeness and agreeableness.
Figure \ref{img:data}(d) confirms such an observation by providing the 15 most frequently elicited emotions by agents' contents.
Although the most represented emotions are \emph{curiosity} and \emph{approval}, it is worth noticing the high volumes of \emph{anger}, \emph{disapproval} and \emph{disappointment} - thus underlining that agents were able to leverage their profiles while generating contents and reacting to peers.

Finally, Figure \ref{img:threads}(a-b) provides examples of agents' generated discussion threads: one focusing on a post written solely by elaborating the interest and profile provided as input for the LLM agent, the other taking as additional input some news extracted by the RSS feed of an online media outlet.
As observed, both threads maintain topical consistency with agents elaborating the discussion, providing different angles and integrating hashtags, mentions and reactions\footnote{The visual mockups were created using \url{https://www.tweetgen.com/}.}.
Moreover, Figure \ref{img:threads}(c) reports the cumulative decreasing distribution of comments received per post (i.e., the discussion thread lengths).
Although most posts receive few comments, it is worth noticing the presence of viral content able to collect hundreds of replies.
Such a trend is also supported by Figure \ref{img:threads}(d) reporting the number of times posts have been suggested by the content recommender systems to agents as inputs for comments/reactions: indeed, algorithmic bias favours a restricted number of contents ($\leq$ 20\%) that were (potentially) able to receive tens of thousands of impressions.

\begin{figure*}[t]
    \centering
    \subfloat[]{\includegraphics[width=0.5\linewidth]{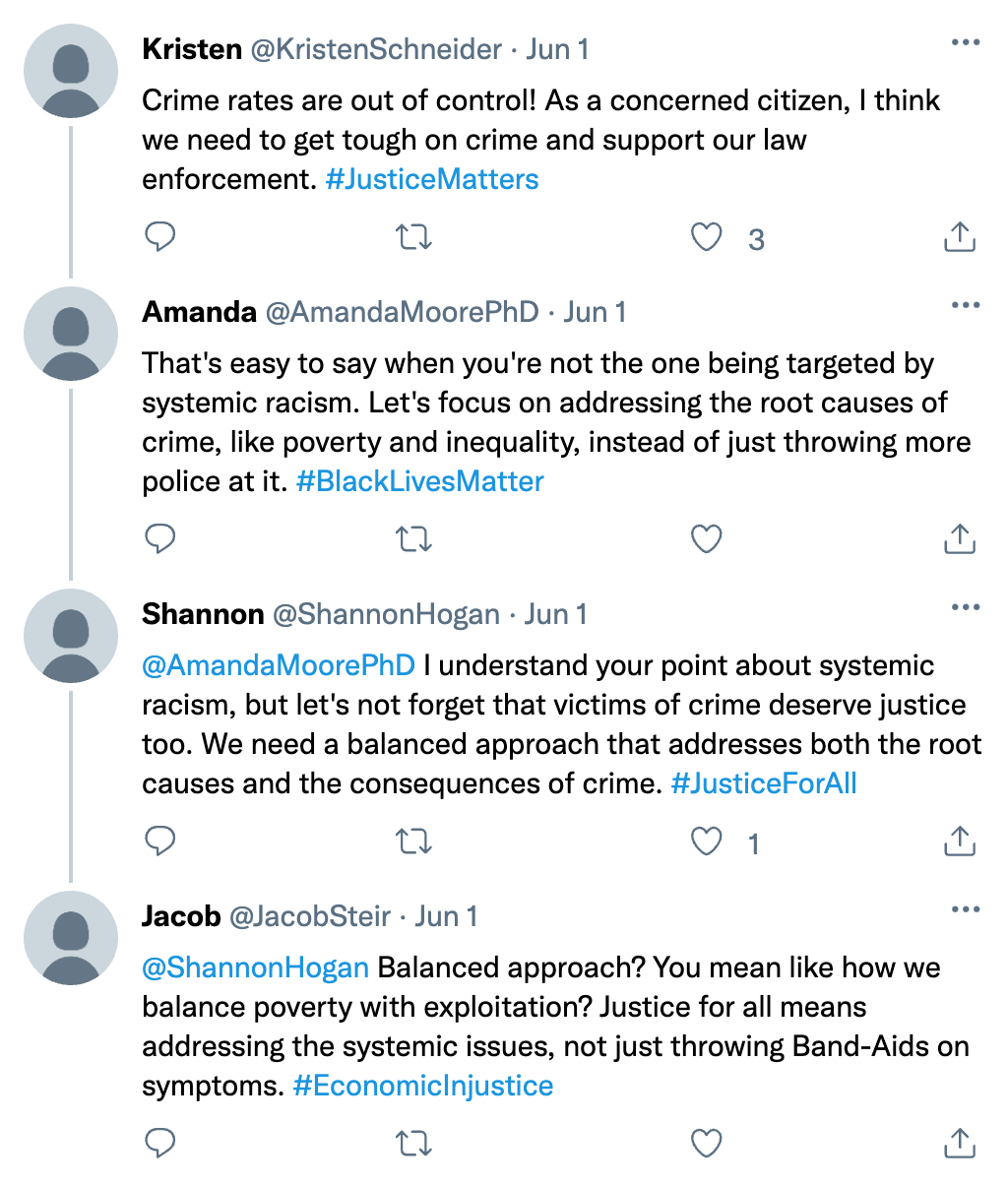}} \qquad \qquad
    \subfloat[]{\includegraphics[width=0.36\linewidth]{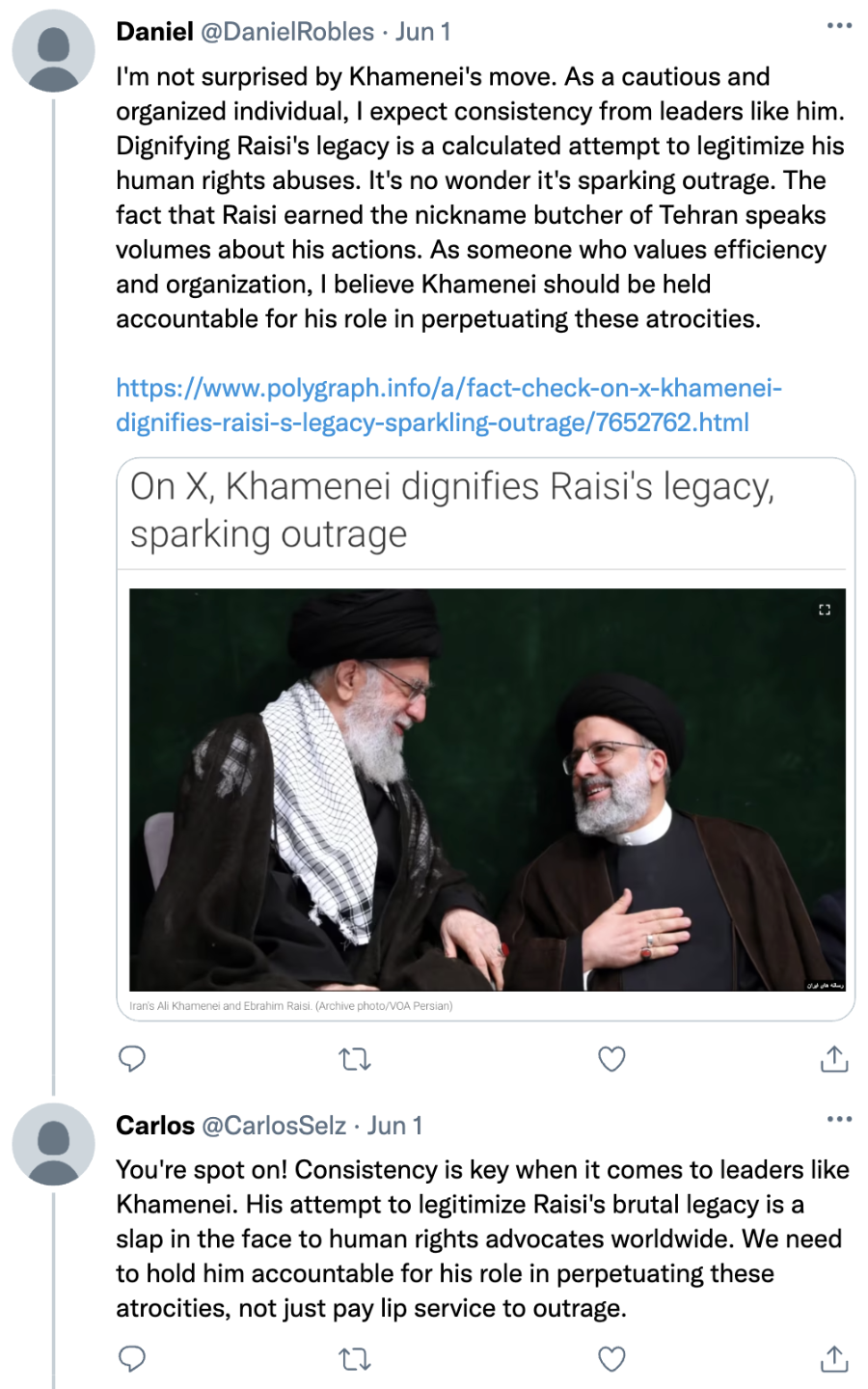}} \\
    \subfloat[]{\includegraphics[width=.5\linewidth]{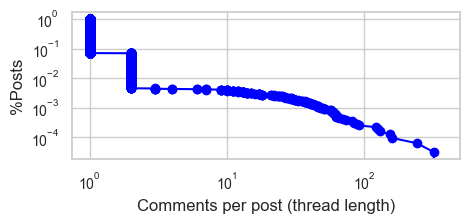}} 
    {\includegraphics[width=.5\linewidth]{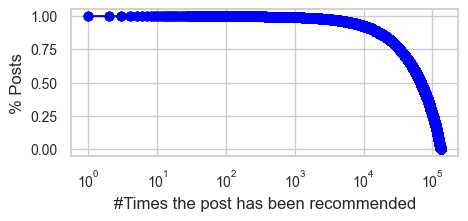}} 
    
    \caption{{\bf Discussion threads.} Examples of (a) LLM-generated discussion thread leveraging agent profiles and interests; (b) Thread based on a piece of news accessed via RSS feeds; (c) Cumulative decreasing distribution of thread lengths - i.e., number of comments per post; (d) CCDF of the number of times a post has been recommended for agents to read.}
    \label{img:threads}
\end{figure*}

\newpage
\section{Leveraging the Y Digital Twin to boost multidisciplinary research}
\label{sec:multid}

In this section, we explore the potential impact of digital twins of online social media platforms, such as {\tt Y}, can have on multiple research fields. 

Although our overview does not provide an exhaustive list of scenarios where these systems can be beneficial, it highlights their importance in supporting multidisciplinary research grounded in ``social" data. 
This is particularly relevant today as easy-to-access online social media data has significantly decreased due to platforms' policies, resulting in a concerning trend requiring more than ever the development of tools and techniques that can provide reliable synthetic data-proxies.

\subsection{Network Science}
Network science is an interdisciplinary domain, focusing on understanding and analyzing the structure, dynamics, and function of complex networks. 
Online social networks have been the focus of research in several directions, such as influence and information spread, community detection and analysis, homophily, polarization, etc. A platform like {\tt Y} offers a remarkable tool for designing controlled experiments and serves as a null model for targeted studies. In the following, we explore some research questions that could benefit from {\tt Y}.

\noindent{\bf Dynamics of/on networks.}
Network structure is well-known to be driven by social aspects, such as preferential attachment (i.e., rich-get-richer effect), or triadic closure (i.e., friends of my friends are my friends). This phenomenon has been observed in real networks\cite{cheng2019socially,huang2014mining}, but one can only study one particular instance of a social network, often incomplete, and without knowing the effect of external factors, e.g., meeting persons outside the network. 
Some recent works\cite{de2023emergence,papachristou2024network} have studied how LLM-powered agents might tend to reproduce these human tendencies, using simple scenarios that focus only on topology. 
With {\tt Y}, one could go much further in that direction, by studying how different parameters might lead to the emergence -- or not -- of this phenomenon. For instance, the rich-get-richer phenomenon is known to be related to a user's exposure to other users' accounts, reputation and activity. 
By modulating these parameters, one could study how emerging topological properties arise from drivers such as post visibility, users' behaviors (consensual vs. confrontational), and similarity of opinions, etc.

Another common topic of research concerns the way information spreads through networks. Information diffusion is relevant for multiple questions, from the diffusion of fake news to social contagion, or the identification of local and global influencers. 
In this context, again, {\tt Y} could be used to explore in a controlled environment the influence of factors such as the network structure, or the role of recommendation algorithms. 
For instance, one could experiment with the factors leading (or not) to a higher level of polarization, from varying the structure of the network (presence of well-separated network communities, presence of large hubs having polarizing opinions) to manipulating the strategies adopted by agents and by recommender systems.

\noindent{\bf Centralities and Network Descriptor Evaluation.}
Popular research topics in network science include developing \textit{network centralities} -- scores describing the role of nodes in a network, e.g., degree centrality, PageRank, Betweenness, etc.-- and \textit{network descriptors} -- scores capturing a property of interest of the networks, e.g., clustering coefficient, coreness, homophily, etc. 
These properties are used in online social networks, for instance, to identify key entities or to evaluate the polarization of the network, or its susceptibility to be manipulated by malicious agents. 
In real networks, it is often difficult to evaluate the relevance of such scores, due to the absence of a ground truth, and the impossibility of replicating experiments. 
With {\tt Y}, one can use controlled experiments and multiple replications. 
For instance, one could be interested in studying how the centrality of an entity is a good predictor of its influence. 
However, it is usually impossible to differentiate the centrality value from other aspects, such as the opinion of the entity, its frequency of activity, or even its exact position in the network. With {\tt Y}, one could keep the same pool of users, but run multiple experiments, in which these entities will occupy different positions in the network. 
This would make it possible to differentiate the role of the network position, comparatively with the behaviour and social strategy of users.

\subsection{Social AI}
The field of Social AI has emerged as a critical area of research at the intersection of artificial intelligence, social psychology, and network science. It aims to understand and model the complex dynamics of human behavior in digital social environments, particularly focusing on the interplay between human users and AI systems. Over the past decade, researchers have increasingly recognized the significant impact that online platforms and their underlying algorithms have on individual behavior, collective decision-making, and societal outcomes \cite{pedreschi2023human, nguyen2020echo}.
A key focus within Social AI research has been the study of information environments in online platforms. Several studies have highlighted how these digital spaces can become "polluted", exacerbating users' cognitive biases and giving rise to concerning phenomena such as the spread of misinformation \cite{hernon1995disinformation}, extreme partisanship, opinion polarization, and the formation of echo chambers \cite{cinelli2021echo, garimella2018quantifying, ge2020understanding, hilbert2018communicating}. 
At the same time, the homophilic mechanisms that drive polluted realities \cite{mcpherson2001birds, newman2003mixing, barbera2015tweeting, kordzadeh2022algorithmic, karimi2023inadequacy, rossetti2021conformity} may have different and potentially beneficial effects when observed in other contexts -- e.g., supportive online groups can form around people who are facing particular challenges offering help and assistance -- suggesting the dual nature of these digital social dynamics.

Central to Social AI research is the concept of Human-AI Co-Evolution, which examines the mutual interactions and feedback loops between users and AI algorithms, shaping behaviors on both sides \cite{pedreschi2023human}. This involves studying how recommendation systems affect content consumption, opinion formation, relationship building, and emotional responses \cite{pappalardo2024survey}. Empirical observational studies \cite{zhou2010impact, hosseinmardi2024causally, yang2023bubbles} provide valuable insights into real-world user behaviors and algorithmic influences. However, they often lack control over external variables and are subject to measurement biases. Empirical controlled studies \cite{ludwig2023divided, bartley2021auditing}, while crucial for establishing causality, face challenges in implementation due to limited access to proprietary platform data and ethical considerations. Simulation studies \cite{sirbu2019algorithmic, pansanella2022modeling, pansanella2023mass} offer a middle ground, allowing for reproducibility and controlled experimentation, but may rely on simplified models that do not fully capture the complexity of real-world behaviors. Notably, there is a lack of controlled simulation studies within social media ecosystems \cite{pappalardo2024survey}, which limits our understanding of how different recommendation algorithms interact with network effects in realistic settings. 

In this context, {\tt Y} emerges as a powerful tool for advancing Social AI research across multiple domains. 
By providing a platform for simulating realistic user interactions under controlled conditions, {\tt Y} addresses several key challenges and opportunities in the field:
{\tt Y} enables researchers to simulate complex social dynamics by manipulating both human behaviors and algorithmic interventions. 
This dual control allows for the evaluation of individual effects on social network topology, user beliefs, and behaviors. 
For example, researchers can model scenarios with varying degrees of social clustering, opinion homogeneity within groups, and algorithmic content exposure. 
By adjusting these parameters across simulations, {\tt Y} facilitates the quantification of their impacts on agent opinions and overall system dynamics.
This fine-grained control over both human and algorithmic factors provides unique insights into the formation, evolution, and potential dissolution of phenomena such as epistemic bubbles, echo chambers, and filter bubbles \cite{nguyen2020echo,nguyen2020cognitive,pariser2011filter}. 
Moreover, {\tt Y}'s flexibility allows for the incorporation of real-world data, enabling more realistic simulations. 
Researchers can introduce elements such as specific discussion topics, diverse personas, influential agents like news outlets, and even simulate misinformation spread and debunking efforts.
In the specific context of supportive information systems, {\tt Y} could offer insights into the effectiveness of various moderation strategies and content curation approaches. 
By simulating different interventions, researchers can assess their impact on user engagement, the quality of discourse, and the overall health of the online community. 
This is particularly relevant for understanding how to foster positive online environments while mitigating potential negative effects of social media interactions.

\subsection{NLP and content analysis}
Since their beginnings, Natural Language Processing and other related fields, such as computational linguistics, have faced significant challenges, especially in domain-specific or multilingual tasks. For example, in cases such as hate speech detection \cite{kovcs2021challenges}, data are still scarce for developing models that automatically recognize linguistic and paralinguistic cues.
Moreover, since language is a complex system composed of both common and uncommon linguistic events, it is almost impossible for every phenomenon to be represented in a sample of textual data, leading to the emergence of data sparsity \cite{guthrie-etal-2006-closer}.
Such a problem has recently been addressed by the development of methods of data augmentation, which can operate at the token level, e.g. by applying a local modification (deletion, swapping or insertion of a token) to a text while preserving its meaning or at the sentence level, i.e. by paraphrasing a given sentence  \cite{chen2023empirical}.
This issue persists for specific tasks, preventing the development of a reliable model of language for the recognition of a particular aspect of language. Furthermore, specific NLP tasks may not be fully solved only by the analysis of textual dimensions from a merely morphological, syntactic or semantic perspective. Still, they could benefit from the interplay with tools offered by other disciplines.

One such task is stance detection, which is usually applied to textual data extracted from social networks \cite{kk2020stance, walker2012thats, mohammad2017stance} to understand the standpoint of a user toward a controversy.
{\tt Y} could ease two issues related to this task. First, it would allow the creation of new datasets or the augmentation of existing datasets, allowing the generation of text data on specific controversies, as one of the issues of developing effective stance detection algorithms is due to the lack of annotated datasets \cite{hardalov2022survey, alturayeif2023systematic}. 
Secondly, it could help in developing stance detection approaches leveraging both network features (e.g., mentions) and natural language processing.
A similar approach could also be useful for argumentation mining, which investigates the argument structure in a given text \cite{Lippi2016argumentation}.
Another common application of NLP is the extraction of emotions and sentiments, another classification task, usually performed through transformer-based models, for which LLMs already demonstrated their strength \cite{krugmann2024survey}.
In this scenario, the text generated by {\tt Y} agents is annotated on the fly by the LLM itself, simplifying and speeding up the annotation process that would otherwise be done by another classification algorithm or another LLM.
{\tt Y} data can be further enhanced with additional information from the texts, such as integrating Named-Entities Recognition (NER) \cite{li2020survey}.
Moving to Natural Language Generation, {\tt Y} can offer a playground in which to explore the potentialities and limitations of LLMs interacting in a social context. 
{\tt Y} could provide data that can be analysed from multiple perspectives, such as the coherence of the text in a long thread of discussion -- to assess LLM's ability to maintain coherence through multiple answers to a post -- and the creativity and reasoning in their responses.

\subsection{Human and Machine Psychology}

While psychology is a vastly affirmed discipline investigating humans' mental states and behavior, machine psychology is a far newer field investigating artificial intelligence within pre-existent or novel cognitive and psychological frameworks \cite{hagendorff2023human,abramski2023cognitive}. 
{\tt Y} can serve as a powerful framework for investigating several psychological patterns, either in humans or in machines.

Social media use can significantly influence human emotions. 
Positive interactions - receiving likes and supportive comments - can boost mood and self-esteem \cite{valkenburg2006}. 
Conversely, negative interactions - cyberbullying or social exclusion - can lead to emotional distress \cite{valkenburg2006}. 
Online platforms can exacerbate negative interactions also because of the so-called disinhibition effect, where people say things online they would not say face-to-face \cite{suler2004}. 
{\tt Y} could be an interesting reference model for detecting disinhibition effects across several topics by analyzing online content, like social media posts, comments, and interactions. By having LLMs produce topic-matched conversations, mimicking threads in real-world online social media, researchers would have access to novel linguistic datasets mapping disinhibited actions as done in \cite{moor2010}, e.g. offensive language, self-disclosure, or trolling, on a vast array of topics. 
Counting differences in disinhibited actions between LLMs and humans engaging in the same topic would provide a language-based, ecological measure of disinhibition levels. In terms of machine psychology, experiments based on the {\tt Y} platform could test whether anonymity and differences between online and offline personas would be enough to reproduce disinhibition effects also in large language models.

Frustration and other negative feelings can derive from social media interactions in terms of unfair personal comparisons. Festinger's social comparison theory posits that individuals evaluate themselves based on comparisons with others \cite{festinger1954}, at least in face-to-face interactions. However, Vogel and colleagues \cite{vogel2014} found that also social media use is associated with increased self-comparison, often leading to negative self-evaluations. Users frequently compare their personal lives to the curated, often idealized, presentations of others, and this results in feelings of inadequacy and lower self-esteem \cite{toma2013}. 
The {\tt Y} platform could be used as an intriguing baseline for understanding how more capable LLMs, e.g. with a larger number of parameters and better training processes, reacted to online social comparisons. 
In the presence of LLMs complex enough to reproduce human negative self-evaluations, researchers might use the {\tt Y} platform for testing which cognitive messages or mechanics could safeguard simulated online users from unfair comparisons. 
In this way, the protection mechanisms designed within {\tt Y} could be tested for protecting online human users from excessively negative and unfair self-evaluations. 

Protection mechanisms designed with {\tt Y} could focus on misinformation too. 
Several recent findings indicate that the social re-sharing of online fake content is promoted by individuals with lower educational attainment \cite{guess2019less}, lower cognitive reflection skills \cite{pennycook2019} and higher inattention \cite{pennycook2021psychology}. 
These psychological dimensions all agree with the framework of cognitive load theory \cite{sweller1988}, which posits that excessive demands of information processing can impair decision-making. 
In social media where users are bombarded with information, the re-sharing of inaccurate content might be due to a lack of time and cognitive resources rather than to an inability to fully understand simple portions of knowledge \cite{pennycook2021psychology,guess2019less}. 
Advanced LLMs would be potentially free from these cognitive limitations. 
Once simulated within the {\tt Y} framework, novel experiments might test how the interplay between LLMs' beliefs and the content of fake news would affect misinformation spread in case cognitive load limitations were mostly resolved. Within {\tt Y}, investigating which type of misinformation content was still promoted by LLMs would provide valuable insights for designing novel LLM-assisted applications fighting misinformation. 
We posit that LLMs present crucial human and non-human psychological biases \cite{abramski2023cognitive}, which translate into LLMs' limitations in fully grasping or understanding misinformation content. 
Hence, despite being mostly free from cognitive overloads, LLMs might still potentially exacerbate misinformation spreading because of a lack of deeper reasoning. 
However, using {\tt Y} would be highly beneficial in understanding under which scenarios did LLMs fail in discerning and stopping the re-sharing of misinformation content. In the presence of such knowledge, future human-AI systems might employ LLMs only in tested scenarios to help humans be less exposed to misinformation cascades.

\subsection{Communication Science}
Social media enables asynchronous communication, where users can engage in conversations without being simultaneously present.
The hyperpersonal communication model \cite{walther1996} suggests that online interactions can sometimes exceed face-to-face communication in intimacy and intensity, since users have more time to select and produce their personal content compared to in-person conversations. 
An example could be a thoughtful comment left on a friend’s post, which might be more considered than a spontaneous remark. 
Recent research has confirmed that users perform a selective self-presentation in online social media, posting only their best photos and achievements and thus creating an unrealistically filtered portrayal of their lives \cite{toma2013}. 
These acts are performed by users to enhance their self-esteem and appeal. 
At the same time, users can end up feeling anxious about maintaining overly idealized personas. 
These elements can ultimately lead to exceedingly and unrealistically positive feedback loops, misrepresentations and idealized perceptions of others \cite{walther1996}. 

A platform like {\tt Y} could produce intriguing simulations where users would not engage in hyperpersonal communication despite being online. 
Crucially, LLMs would react immediately to online content, thus producing continuous and non-delayed conversations. 
This difference could lead to LLMs being less prone to commit to selective self-presentation, potentially giving rise to fewer idealizations and misrepresentations than human users when being in similar online scenarios. 
To this aim, LLMs should be instructed to avoid a positive bias in their language, which could still also lead to shallower conversations compared to humans, as the latter would have more time to produce deeper conversations, according to \cite{walther1996}. 

The presence and influence of hyperpersonal communication and its implications could also be tested in online dating social media (like those tested in \cite{toma2013}) reconstructed via the {\tt Y} platform. 
Y-based digital twins would have LLMs rather than people engaging in conversations aimed at finding potential dating matches. These simulations would open to unprecedented investigations of what kind of features and emotional reactions would entail LLMs compared to self-selective humans feeling the pressure of dealing with idealized perceptions of others.

\newpage
\section{Conclusion and Future Works}
\label{sec:conclusions}
In this work, we introduced {\tt Y Social} an LLM-powered Online Social Media platform Digital Twin.

{\tt Y} is designed to allow the description and execution of complex agent-based social simulations avoiding mechanistic/deterministic solutions and allowing researchers and practitioners to perform controlled what-if scenario testing.

To overcome existing solutions' limitations, {\tt Y} focuses on simulating social interactions and content generation at once, making those two dimensions naturally intertwined in the data generation process.

As future work, we plan to apply the developed system to multiple, multidisciplinary, research endeavours - some of which are sketched in Section \ref{sec:multid} - and to extend the platform to include additional simulation layers and features - e.g., including specific agent-roles (such moderator, bot, gatekeeper\dots) and expanding the actions available to agents.
Moreover, we plan to develop three different frontends for the simulation platform: (i) a social media-like web application to showcase the simulation and allow humans to interact with LLM agents - thus implementing a hybrid simulation environment; (ii) an interactive and easy to use simulation configuration tools to make zero-code the simulation (and agent population) configuration and deployment; (iii) a data analysis dashboard to allow online explorative characterization of the simulation generated data.

{\tt Y} along with its future extensions, evolution, simulation data and related configuration ``recipes" are  publicly available on the official GitHub organization page\footnote{{\tt Y} GitHub: \url{https://github.com/YSocialTwin}}, website \footnote{{\tt Y} Website: \url{https://YSocialTwin.github.io}} and within the SoBigData EU Research Infrastructure\footnote{SoBigData RI: \url{http://sobigdata.eu/}}.

\subsection*{Acknowledgments}
This work is supported by (i) the European Union – Horizon 2020 Program under the scheme “INFRAIA-01-2018-2019 – Integrating Activities for Advanced Communities”, Grant Agreement n.871042, “SoBigData++: European Integrated Infrastructure for Social Mining and Big Data Analytics” (\url{http://www.sobigdata.eu}); (ii) SoBigData.it
which receives funding from the European Union – NextGenerationEU – National Recovery and Resilience Plan (Piano Nazionale di Ripresa e Resilienza, PNRR) – Project: “SoBigData.it – Strengthening the Italian RI for Social Mining and Big Data Analytics” – Prot. IR0000013 – Avviso n. 3264 del 28/12/2021; (iii) EU NextGenerationEU programme under the funding schemes PNRR-PE-AI FAIR (Future Artificial Intelligence Research); (iv) SAGE Concept Grants.

\bibliographystyle{abbrv}
\bibliography{references}

\end{document}